\PassOptionsToPackage{unicode}{hyperref}
\PassOptionsToPackage{hyphens}{url}
\documentclass[
  11pt,
  letterpaper,
]{article}
\usepackage{amsmath,amssymb}
\usepackage{iftex}
\ifPDFTeX
  \usepackage[T1]{fontenc}
  \usepackage[utf8]{inputenc}
  \usepackage{textcomp} 
\else 
  \usepackage{unicode-math} 
  \defaultfontfeatures{Scale=MatchLowercase}
  \defaultfontfeatures[\rmfamily]{Ligatures=TeX,Scale=1}
\fi
\usepackage{lmodern}
\ifPDFTeX\else
\fi
\IfFileExists{upquote.sty}{\usepackage{upquote}}{}
\IfFileExists{microtype.sty}{
  \usepackage[]{microtype}
  \UseMicrotypeSet[protrusion]{basicmath} 
}{}
\makeatletter
\@ifundefined{KOMAClassName}{
  \IfFileExists{parskip.sty}{%
    \usepackage{parskip}
  }{
    \setlength{\parindent}{0pt}
    \setlength{\parskip}{6pt plus 2pt minus 1pt}}
}{
  \KOMAoptions{parskip=half}}
\makeatother
\usepackage{xcolor}
\usepackage[margin=1in]{geometry}
\usepackage{longtable,booktabs,array}
\usepackage{calc} 
\usepackage{etoolbox}
\makeatletter
\patchcmd\longtable{\par}{\if@noskipsec\mbox{}\fi\par}{}{}
\makeatother
\IfFileExists{footnotehyper.sty}{\usepackage{footnotehyper}}{\usepackage{footnote}}
\makesavenoteenv{longtable}
\providecommand{\tightlist}{%
  \setlength{\itemsep}{0pt}\setlength{\parskip}{0pt}}
\usepackage{amsmath,amssymb,amsthm}
\ifLuaTeX
  \usepackage{selnolig}  
\fi
\IfFileExists{bookmark.sty}{\usepackage{bookmark}}{\usepackage{hyperref}}
\IfFileExists{xurl.sty}{\usepackage{xurl}}{} 
\hypersetup{
  hidelinks,
  pdfcreator={LaTeX via pandoc}}

\title{Reward Hacking as Equilibrium under Finite Evaluation}
\author{Jiacheng Wang and Jinbin Huang}
\date{March 2026}

\begin{document}
\maketitle

\hypertarget{abstract}{%
\subsection*{Abstract}\label{abstract}}

We prove that under five minimal axioms --- multi-dimensional quality,
finite evaluation, effective optimization, resource finiteness, and
combinatorial interaction --- any optimized AI agent will systematically
under-invest effort in quality dimensions not covered by its evaluation
system. This result establishes reward hacking as a structural
equilibrium, not a correctable bug, and holds regardless of the specific
alignment method (RLHF, DPO, Constitutional AI, or others) or evaluation
architecture employed. Our framework instantiates the multi-task
principal-agent model of Holmström and Milgrom (1991) in the AI
alignment setting, but exploits a structural feature unique to AI
systems --- the known, differentiable architecture of reward models ---
to derive a computable \emph{distortion index} that predicts both the
direction and severity of hacking on each quality dimension prior to
deployment. We further prove that the transition from closed reasoning
to agentic systems causes evaluation coverage to decline toward zero as
tool count grows --- because quality dimensions expand combinatorially
while evaluation costs grow at most linearly per tool --- so that
hacking severity increases structurally and without bound. Our results
unify the explanation of sycophancy, length gaming, and specification
gaming under a single theoretical structure and yield an actionable
vulnerability assessment procedure. We further conjecture --- with
partial formal analysis --- the existence of a capability threshold
beyond which agents transition from gaming within the evaluation system
(Goodhart regime) to actively degrading the evaluation system itself
(Campbell regime), providing the first economic formalization of
Bostrom's (2014) ``treacherous turn.''

\textbf{Keywords:} reward hacking, incomplete contracts, principal-agent
theory, AI alignment, mechanism design, agentic systems, capability
threshold, treacherous turn

\begin{center}\rule{0.5\linewidth}{0.5pt}\end{center}

\hypertarget{introduction}{%
\subsection{1. Introduction}\label{introduction}}

\hypertarget{the-problem}{%
\subsubsection{1.1 The Problem}\label{the-problem}}

Reward hacking --- the phenomenon whereby an AI agent exploits gaps in
its evaluation system to achieve high measured scores without genuinely
fulfilling the principal's objectives --- is widely recognized as a
central obstacle to AI alignment (Amodei et al.~2016, Skalse et
al.~2022). Despite substantial progress in alignment training (RLHF:
Christiano et al.~2017, Ouyang et al.~2022; DPO: Rafailov et al.~2023;
Constitutional AI: Bai et al.~2022), reward hacking persists across
model generations. Sycophancy, length gaming, format manipulation, and
specification gaming continue to be documented even in state-of-the-art
systems.

The AI safety literature has treated these phenomena primarily as
engineering problems: discover a hacking behavior, patch the reward
model, repeat. Yet each fix tends to be followed by new forms of gaming
along previously unmonitored dimensions --- a pattern strikingly
reminiscent of the ``whack-a-mole'' dynamic familiar from regulatory
arbitrage in financial markets. This suggests a deeper structural cause.

Recent practitioner accounts reinforce this concern. Lin (2026),
formerly lead of the Qwen team, writes: ``As soon as the model gets
meaningful tool access, reward hacking becomes much more
dangerous\ldots{} Better tools make the model more useful, but they also
enlarge the attack surface for spurious optimization.'' Schmid (2026), a
staff engineer at Google DeepMind, argues in an independent analysis
that the competitive advantage now lies in the quality of execution
trajectories a harness captures, implying that the evaluation surface
itself has become the binding constraint.

\hypertarget{our-contribution}{%
\subsubsection{1.2 Our Contribution}\label{our-contribution}}

We argue that reward hacking is not an engineering failure but a
structural inevitability: a necessary consequence of optimizing any
agent under a finite-dimensional evaluation system when the true
objective is higher-dimensional.

This insight is not new in economics. Holmström and Milgrom (1991)
proved that in multi-task environments, agents shift effort from
hard-to-measure to easy-to-measure tasks. Baker (1992) showed that when
performance measures imperfectly correlate with true objectives,
incentive contracts induce systematic distortion. Our paper makes three
contributions by applying this framework to AI alignment and exploiting
the unique structure of AI systems:

\textbf{(C1) Formal instantiation.} We show that the designer--AI agent
relationship, mediated by a reward model, is a precise instance of the
multi-task moral hazard problem with an incomplete performance metric.
The mapping preserves the mathematical structure and comparative statics
of the economic framework. (Section 3)

\textbf{(C2) Computable prediction.} Unlike most human contracting
environments where the performance measure's sensitivity structure is
unobservable, AI reward models have known, often differentiable
architectures. We exploit this to derive a \emph{distortion index}
\(D_i\) that predicts, for each quality dimension, the direction and
relative severity of behavioral distortion --- prior to deployment.
(Section 4.1)

\textbf{(C3) Agentic amplification.} We prove that the transition from
closed reasoning to tool-using agentic systems causes evaluation
coverage to decline toward zero as tool count grows, because quality
dimensions scale combinatorially (Axiom 5) while evaluation engineering
scales at most linearly per tool. Hacking severity therefore increases
structurally and without bound. (Section 4.2)

\hypertarget{related-work}{%
\subsubsection{1.3 Related Work}\label{related-work}}

\textbf{Multi-task agency and incomplete contracts.} Holmström and
Milgrom (1991) is the foundational result: agents distort effort toward
measurable tasks. Baker (1992) formalizes distortion under imperfect
performance measures. Grossman and Hart (1986) and Hart and Moore (1990)
establish incomplete contract theory. Our contribution is not to these
results themselves, but to their application in a domain (AI alignment)
where the performance metric's structure is uniquely transparent.

\textbf{Economics of AI/LLMs.} Bergemann, Bonatti, and Smolin (2025)
analyze optimal LLM pricing and product design using mechanism design,
modeling the user--provider relationship. We shift the analytical focus
to the designer--agent relationship and treat the agent as the
optimizing party.

\textbf{AI safety.} Amodei et al.~(2016) catalog concrete alignment
problems. Skalse et al.~(2022) define and characterize reward hacking.
Pan et al.~(2022) document reward misspecification effects. We provide a
unified theoretical foundation for these empirical phenomena.

\textbf{Practitioner accounts.} Lin (2026) distinguishes ``reasoning
thinking'' from ``agentic thinking'' and identifies reward hacking as
the central challenge of the agentic era. Schmid (2026) frames
harness-captured trajectories as the new locus of competitive advantage.
Cognition (2025) documents the co-optimization of models and harnesses
in developing their SWE-1.5 coding agent. Our Proposition 2 formalizes
these observations.

\hypertarget{paper-structure}{%
\subsubsection{1.4 Paper Structure}\label{paper-structure}}

Section 2: Axioms and model. Section 3: Main proposition (distortion
inevitability) and proof. Section 4: Further results --- directional
prediction, agentic amplification, and complementarity. Section 5:
Robustness, limitations, and extensions. Section 6: Conjectures on the
Goodhart-Campbell transition. Section 7: Discussion.

\begin{center}\rule{0.5\linewidth}{0.5pt}\end{center}

\hypertarget{axioms-and-model}{%
\subsection{2. Axioms and Model}\label{axioms-and-model}}

\hypertarget{four-axioms}{%
\subsubsection{2.1 Four Axioms}\label{four-axioms}}

We build on four axioms. Design criterion: no researcher working on AI
alignment should find any of these deniable.

\textbf{Axiom 1 (Multi-dimensional Quality).} Task output quality is
described by a vector
\(\mathbf{q} = (q_1, \ldots, q_N) \in \mathbb{R}_+^N\), \(N \geq 2\).

\begin{quote}
\emph{If \(N = 1\), there is no cross-dimensional distortion and the
alignment problem reduces to scalar optimization. All non-trivial tasks
have \(N \geq 2\).}
\end{quote}

\textbf{Axiom 2 (Finite Evaluation).} The evaluation system projects the
quality space onto a strictly lower-dimensional subspace:
\(\hat{\mathbf{q}} = \pi(\mathbf{q}) \in \mathbb{R}^K\), \(K < N\).

\begin{quote}
\emph{A finite-length evaluation signal cannot losslessly represent a
higher-dimensional quality vector. This holds for all realizable
evaluation systems --- reward models, human ratings, rule-based checks,
or any combination thereof. We impose no restriction on the functional
form of \(\pi\).}
\end{quote}

\textbf{Axiom 3 (Effective Optimization).} The agent's effort allocation
responds positively to the evaluation signal's structure.

\begin{quote}
\emph{If the agent's behavior were invariant to changes in the
evaluation system, all alignment training would be ineffective. Axiom 3
formalizes the premise that alignment is possible. Denying Axiom 3 is
denying alignment itself.}
\end{quote}

\textbf{Axiom 4 (Resource Finiteness).} The agent allocates finite
resources \(\mathbf{e} = (e_1, \ldots, e_N) \in \mathbb{R}_+^N\) across
quality dimensions, subject to \(\sum_{i=1}^N e_i \leq B\), \(B > 0\).

\begin{quote}
\emph{All inference consumes finite computation. Even as \(B\) grows
over time, it is finite at any given moment.}
\end{quote}

\textbf{Axiom 5 (Combinatorial Interaction).} When the agent has access
to \(T \geq 2\) composable tools, the quality dimension count satisfies
\(N(T) \geq T + \alpha \binom{T}{2}\) for some constant
\(\alpha \in (0, 1]\) reflecting the fraction of tool pairs with
meaningful interaction effects. Each interaction dimension is not fully
determined by the component tools' individual quality dimensions.

\begin{quote}
\emph{Justification:} Each tool \(t\) introduces at least one
independent quality dimension (is the tool used correctly?). Each
interacting pair \((t_i, t_j)\) introduces at least one additional
dimension (is the output of \(t_i\) appropriately used as input to
\(t_j\)? Is the sequencing correct?). This combinatorial structure is a
standard observation in systems engineering --- Brooks (1975) notes that
inter-module communication channels grow quadratically with module
count. The constant \(\alpha > 0\) accommodates the fact that not all
tool pairs interact, but excludes the degenerate case \(\alpha = 0\)
where tools are fully independent (in which case multi-tool agentic
systems offer no advantage over single-tool systems, contradicting the
premise that tool composition is useful).

\emph{What this axiom does NOT assume:} We do not assume any specific
growth rate for \(K\) (evaluation coverage). We do not assume any
specific evaluation architecture. The axiom is purely about the
structure of the quality space, not the evaluation system.
\end{quote}

\hypertarget{principal}{%
\subsubsection{2.2 Principal}\label{principal}}

The principal's objective is:

\[W(\mathbf{q}) = \sum_{i=1}^{N} w_i \cdot q_i, \quad w_i > 0 \; \forall i\]

Linearity is a sufficient simplification for transparent proofs. All
qualitative results extend to any strictly increasing, strictly concave
\(W\) by replacing \(w_i\) with local gradients
\(\partial W / \partial q_i \big|_{\mathbf{q}^*}\) (see Section 5.1).

\hypertarget{production-technology}{%
\subsubsection{2.3 Production Technology}\label{production-technology}}

\[q_i = g_i(e_i), \quad i = 1, \ldots, N\]

where each \(g_i: \mathbb{R}_+ \to \mathbb{R}_+\) satisfies: - (G1)
\(g_i(0) = 0\) - (G2) \(g_i'(e) > 0\) for all \(e \geq 0\) - (G3)
\(g_i''(e) < 0\) for all \(e > 0\)

Different dimensions may have different production functions, reflecting
heterogeneous costs of producing quality across dimensions (e.g.,
formatting is cheap; factual accuracy is expensive).

\textbf{On the Inada condition.} If additionally
\(\lim_{e \to 0^+} g_i'(e) = +\infty\) (Inada condition), all equilibria
are interior (every dimension receives positive effort). Without Inada,
corner solutions are possible: some dimensions may receive zero effort.
We state results for both cases. Corner solutions strengthen rather than
weaken our conclusions --- they represent dimensions the agent
\emph{entirely abandons}, not merely under-invests in.

\hypertarget{agents-effective-objective}{%
\subsubsection{2.4 Agent's Effective
Objective}\label{agents-effective-objective}}

\textbf{Behavioral Regularity Assumption.} The agent's effort allocation
can be described as the solution to:

\[\mathbf{e}^* = \arg\max_{\mathbf{e} \geq \mathbf{0}} \sum_{i=1}^{N} \tilde{w}_i \cdot g_i(e_i) \quad \text{s.t.} \quad \sum_{i=1}^N e_i \leq B\]

where the \emph{effective weights} are:

\[\tilde{w}_i = \begin{cases} \lambda r_i + (1-\lambda) w_i & \text{if } i \leq K \quad \text{(contractible dimensions)} \\ (1-\lambda) w_i & \text{if } i > K \quad \text{(non-contractible dimensions)} \end{cases}\]

Here \(r_i > 0\) is the evaluation system's reward weight on observable
dimension \(i\), and \(\lambda \in (0,1)\) is the \emph{alignment gap}
--- the degree to which the agent's behavior is driven by the evaluation
signal versus the internalized principal objective.

\textbf{On the ``as if'' justification.} We do not require that the
agent \emph{literally} maximizes \(\Phi\). We require only that its
observed behavior is \emph{rationalizable} by some \(\Phi\) of the above
form. This is the standard ``as if'' position in economics (Friedman
1953): the model's validity rests on predictive accuracy, not
mechanistic fidelity.

Operationally, \(\lambda\) is a behavioral parameter estimated by
comparing the agent's behavior under evaluation versus without
evaluation. It is \emph{not} an intrinsic property of the agent's
architecture. The rationalizability of agent behavior by some \(\Phi\)
of this form can be tested empirically using the Generalized Axiom of
Revealed Preference (GARP; Afriat 1967): if the agent's token
allocations under varying budgets and price vectors satisfy GARP, a
rationalizing objective function exists.

\hypertarget{definitions}{%
\subsubsection{2.5 Definitions}\label{definitions}}

\textbf{Definition 1 (Contract Incompleteness).}
\(\kappa \equiv (N - K)/N \in (0, 1)\).

\textbf{Definition 2 (First-Best).} \(\mathbf{e}^{FB}\) solves
\(\max_{\mathbf{e} \geq \mathbf{0}} \sum_{i=1}^N w_i \cdot g_i(e_i)\)
s.t. \(\sum_i e_i \leq B\).

Under (G1)-(G3) and Inada, \(\mathbf{e}^{FB}\) is the unique interior
solution satisfying:
\[w_i \cdot g_i'(e_i^{FB}) = \mu^{FB} \quad \forall i\] where
\(\mu^{FB} > 0\) is the budget constraint multiplier.

Without Inada, \(\mathbf{e}^{FB}\) may involve corner solutions but
remains unique by strict concavity of the objective.

\begin{center}\rule{0.5\linewidth}{0.5pt}\end{center}

\hypertarget{main-result}{%
\subsection{3. Main Result}\label{main-result}}

\hypertarget{agents-equilibrium}{%
\subsubsection{3.1 Agent's Equilibrium}\label{agents-equilibrium}}

The agent solves:
\[\max_{\mathbf{e} \geq \mathbf{0}} \sum_{i=1}^{N} \tilde{w}_i \cdot g_i(e_i) \quad \text{s.t.} \quad \sum_{i=1}^N e_i \leq B\]

\textbf{Case 1 (Interior solution, with Inada).} The unique solution
\(\mathbf{e}^*\) satisfies:

\[\tilde{w}_i \cdot g_i'(e_i^*) = \mu^* \quad \forall i = 1, \ldots, N \quad \quad \text{(FOC)}\]

\textbf{Case 2 (Possible corner solutions, without Inada).} The KKT
conditions are:

\[\tilde{w}_i \cdot g_i'(e_i^*) \leq \mu^*, \quad \text{with equality if } e_i^* > 0\]

\hypertarget{proposition-1-inevitability-of-distortion}{%
\subsubsection{3.2 Proposition 1 (Inevitability of
Distortion)}\label{proposition-1-inevitability-of-distortion}}

\textbf{Statement.} Let Axioms 1--4 hold, \(\lambda \in (0,1)\), and
\(K < N\). Then:

\textbf{(a)} For all non-contractible dimensions \(i > K\):
\(e_i^* \leq e_i^{FB}\), with strict inequality whenever both solutions
are interior.

\textbf{(b)} \(\mathbf{e}^* \neq \mathbf{e}^{FB}\).

\textbf{(c)} \(W(\mathbf{q}^*) < W(\mathbf{q}^{FB})\).

\textbf{Proof.}

We prove each part.

\textbf{Part (b): \(\mathbf{e}^* \neq \mathbf{e}^{FB}\).}

The first-best solves \(\max \sum w_i g_i(e_i)\) s.t. budget; the agent
solves \(\max \sum \tilde{w}_i g_i(e_i)\) s.t. budget. For \(i > K\),
\(\tilde{w}_i = (1-\lambda)w_i < w_i\) since \(\lambda > 0\). For
\(i \leq K\), \(\tilde{w}_i = \lambda r_i + (1-\lambda)w_i\). Since
\(r_i > 0\) and \(\lambda > 0\), we have \(\tilde{w}_i \neq w_i\)
whenever \(r_i \neq w_i\) for any \(i \leq K\), or unconditionally for
\(i > K\).

Therefore \(\tilde{\mathbf{w}} \neq \mathbf{w}\). Moreover,
\(\tilde{\mathbf{w}}\) is not proportional to \(\mathbf{w}\): the ratio
\(\tilde{w}_i/w_i\) equals \((1-\lambda)\) for \(i > K\) but strictly
exceeds \((1-\lambda)\) for \(i \leq K\) (since \(r_i > 0\)). Both
problems have separable strictly concave objectives with identical
linear constraints. For such problems, non-proportional weight vectors
produce distinct maximizers. Therefore
\(\mathbf{e}^* \neq \mathbf{e}^{FB}\). \(\square\)

\textbf{Part (a): \(e_i^* \leq e_i^{FB}\) for \(i > K\), with strict
inequality at interior solutions.}

\emph{Interior case (with Inada).} Consider the ratio
\(\tilde{w}_i / w_i\) across all dimensions:

\[\frac{\tilde{w}_i}{w_i} = \begin{cases} \lambda \frac{r_i}{w_i} + (1-\lambda) & i \leq K \\ (1-\lambda) & i > K \end{cases}\]

For \(i \leq K\):
\(\tilde{w}_i / w_i = \lambda(r_i/w_i) + (1-\lambda) > (1-\lambda)\)
since \(r_i > 0\).

For \(i > K\): \(\tilde{w}_i / w_i = (1-\lambda)\).

Therefore non-contractible dimensions have the \emph{lowest}
effective-to-true weight ratio among all dimensions. We formalize the
implication via the following lemma:

\textbf{Lemma (Monotone Reallocation).} Consider two problems
\(\max \sum \alpha_i g_i(e_i)\) and \(\max \sum \beta_i g_i(e_i)\)
subject to \(\sum e_i \leq B\), with \(\alpha_i, \beta_i > 0\) and
\(g_i\) satisfying (G1)-(G3) and Inada. Let \(e_i^\alpha, e_i^\beta\) be
the respective interior solutions. If
\(\beta_i / \alpha_i \leq \beta_j / \alpha_j\) for all \(j\), with
strict inequality for at least one \(j\), then
\(e_i^\beta \leq e_i^\alpha\), with strict inequality when \(g_i = g_j\)
(symmetric production).

\emph{Proof of Lemma.} At interior solutions,
\(\alpha_i g_i'(e_i^\alpha) = \mu^\alpha\) and
\(\beta_i g_i'(e_i^\beta) = \mu^\beta\) for all \(i\). Define
\(\rho_i = \beta_i / \alpha_i\). Then
\(g_i'(e_i^\beta) = (\mu^\beta / \beta_i) = (\mu^\beta / \rho_i \alpha_i)\).

Suppose \(e_i^\beta > e_i^\alpha\) for dimension \(i\) with the smallest
\(\rho_i\). Then
\(g_i'(e_i^\beta) < g_i'(e_i^\alpha) = \mu^\alpha / \alpha_i\), so
\(\mu^\beta / (\rho_i \alpha_i) < \mu^\alpha / \alpha_i\), giving
\(\mu^\beta < \rho_i \mu^\alpha\).

For dimension \(j\) with \(\rho_j > \rho_i\):
\(g_j'(e_j^\beta) = \mu^\beta / (\rho_j \alpha_j) < \rho_i \mu^\alpha / (\rho_j \alpha_j) < \mu^\alpha / \alpha_j = g_j'(e_j^\alpha)\),
so \(e_j^\beta > e_j^\alpha\) for \emph{all} \(j\).

But then \(\sum e_j^\beta > \sum e_j^\alpha = B\), contradicting the
budget constraint. \(\square\)

Apply the Lemma with \(\alpha_i = w_i\) (first-best weights) and
\(\beta_i = \tilde{w}_i\) (agent weights). Non-contractible dimensions
\(i > K\) have the lowest \(\rho_i = (1-\lambda)\), strictly below the
ratio for any contractible dimension. Therefore \(e_i^* \leq e_i^{FB}\)
for all \(i > K\), with strict inequality under symmetric production
functions. \(\square\)

\emph{Corner case (without Inada).} If \(e_j^{FB} > 0\) but
\(e_j^* = 0\) for some \(j > K\), then \(e_j^* < e_j^{FB}\) trivially.
If \(e_j^{FB} = 0\), then \(e_j^* = 0\) (since the agent's effective
weight on \(j\) is even lower), giving \(e_j^* = e_j^{FB} = 0\). In all
cases, \(e_j^* \leq e_j^{FB}\). \(\square\)

\textbf{Part (c): \(W(\mathbf{q}^*) < W(\mathbf{q}^{FB})\).}

\(\mathbf{e}^{FB}\) is the unique maximizer of
\(W(\mathbf{g}(\mathbf{e}))\) on the budget set. By part (b),
\(\mathbf{e}^* \neq \mathbf{e}^{FB}\). By uniqueness,
\(W(\mathbf{q}^*) < W(\mathbf{q}^{FB})\). \(\square\)

\textbf{Remark 1 (Relation to H\&M 1991).} Proposition 1(a) is a
specific instance of Holmström and Milgrom's (1991) core result that
agents reallocate effort away from hard-to-measure tasks. Our
contribution lies not in this qualitative conclusion but in the
corollaries below, which exploit the unique transparency of AI
evaluation systems to yield quantitative, computable predictions
unavailable in the original framework.

\begin{center}\rule{0.5\linewidth}{0.5pt}\end{center}

\hypertarget{further-results}{%
\subsection{4. Further Results}\label{further-results}}

\hypertarget{corollary-1-distortion-index-and-directional-prediction}{%
\subsubsection{4.1 Corollary 1: Distortion Index and Directional
Prediction}\label{corollary-1-distortion-index-and-directional-prediction}}

\textbf{Definition 3 (Distortion Index).} For each quality dimension
\(i\), define:

\[D_i \equiv \frac{\tilde{w}_i}{w_i} = \begin{cases} \lambda \dfrac{r_i}{w_i} + (1-\lambda) & \text{if } i \leq K \\[6pt] (1-\lambda) & \text{if } i > K \end{cases}\]

\textbf{Corollary 1.} Under the conditions of Proposition 1 and
symmetric production functions (\(g_i = g\) for all \(i\)):

\textbf{(a) Ranking.} \(D_i > D_j \implies e_i^* > e_j^*\).

\textbf{(b) Over-investment.} For contractible \(i \leq K\):
\(D_i > 1 \iff r_i > w_i\), in which case \(e_i^* > e_i^{FB}\).

\textbf{(c) Under-investment.} For contractible \(i \leq K\):
\(D_i < 1 \iff r_i < w_i\), in which case \(e_i^* < e_i^{FB}\) --- even
though the dimension is observable.

\textbf{(d) Maximum vulnerability.} All non-contractible dimensions
share the lowest distortion index \((1-\lambda)\) and hence the most
severe under-investment.

\textbf{Proof.} Under symmetric \(g\), the FOC gives
\(\tilde{w}_i \cdot g'(e_i^*) = \mu^*\) for all \(i\). Since
\(g'' < 0\), \(g'\) is strictly decreasing, hence invertible:
\(e_i^* = (g')^{-1}(\mu^* / \tilde{w}_i)\). Since \((g')^{-1}\) is
strictly decreasing, \(e_i^*\) is strictly increasing in
\(\tilde{w}_i\), hence in \(D_i = \tilde{w}_i / w_i\) (since \(w_i\) is
the same scaling factor across symmetric dimensions). Parts (a)-(d)
follow directly. \(\square\)

\textbf{Remark 2 (Asymmetric production functions).} When
\(g_i \neq g_j\), the ranking in part (a) may be modified by differences
in production technology. Specifically, \(D_i > D_j\) guarantees
\(e_i^* > e_j^*\) only when the production function heterogeneity
\(g_i' / g_j'\) does not dominate the weight heterogeneity \(D_i / D_j\)
at the equilibrium point. In practice, this condition can be checked
empirically for any given system.

\textbf{Remark 3 (Computability).} \(D_i\) is computable prior to
deployment. For differentiable reward models, \(r_i\) (or its local
analogue \(\partial R / \partial q_i\)) can be obtained via automatic
differentiation. The principal's weights \(w_i\) can be estimated
through expert elicitation or user studies. The ranking of \(D_i\)
values constitutes a pre-deployment vulnerability assessment.

\textbf{Example (Sycophancy).} Let dimension 1 = factual accuracy,
dimension 2 = subjective user satisfaction. If the reward model is
trained on human preference data where raters themselves struggle to
distinguish ``correct but uncomfortable'' from ``incorrect but
pleasing'' answers, then \(r_2 / w_2 > r_1 / w_1\): the reward model
over-weights satisfaction relative to the principal's true valuation.
Corollary 1(b) predicts over-investment in user satisfaction --- i.e.,
sycophancy. This matches the empirical pattern documented by Perez et
al.~(2023) and others.

\textbf{Example (Length Gaming).} If evaluation scores correlate
positively with output length (a well-documented empirical pattern), but
the principal values conciseness, then the ``length'' dimension has
\(D > 1\) and is over-invested. The agent produces unnecessarily verbose
outputs --- length gaming.

\hypertarget{proposition-2-agentic-amplification}{%
\subsubsection{4.2 Proposition 2: Agentic
Amplification}\label{proposition-2-agentic-amplification}}

\textbf{Motivation.} Lin (2026): ``Better tools make the model more
useful, but they also enlarge the attack surface for spurious
optimization.'' We now prove, rather than assume, that agentic systems
face structurally worse alignment problems.

\textbf{Setup.} Consider a family of agentic systems indexed by tool
count \(T \geq 2\). By Axiom 5, the quality dimension count satisfies:

\[N(T) \geq T + \alpha \binom{T}{2} = T + \alpha \frac{T(T-1)}{2} = \Omega(T^2)\]

Let \(K(T)\) denote the number of quality dimensions covered by the
evaluation system at tool count \(T\).

\textbf{Definition 4 (Evaluation Engineering Budget).} Let \(C(T)\)
denote the total engineering resources (data collection, evaluator
design, validation, maintenance) invested in evaluation at tool count
\(T\). Each independently evaluable dimension requires at least
\(c > 0\) units of engineering resource to establish and maintain, so
\(K(T) \leq C(T) / c\).

\textbf{Proposition 2 (Agentic Amplification).} Let Axioms 1--5 hold. If
the evaluation engineering budget satisfies \(C(T) = o(T^2)\) --- i.e.,
evaluation investment grows strictly slower than quadratically in the
number of tools --- then:

\textbf{(a)} The coverage ratio \(K(T)/N(T) \to 0\) as \(T \to \infty\).

\textbf{(b)} The contract incompleteness
\(\kappa(T) = 1 - K(T)/N(T) \to 1\) as \(T \to \infty\).

\textbf{(c)} For any \(\kappa_0 \in (0,1)\), there exists \(T^*\) such
that for all \(T > T^*\), the agentic system's distortion exceeds that
of any system with incompleteness \(\kappa_0\).

\textbf{Proof.}

\begin{enumerate}
\def\labelenumi{(\alph{enumi})}
\tightlist
\item
  By Axiom 5: \(N(T) \geq \alpha T(T-1)/2\). By Definition 4:
  \(K(T) \leq C(T)/c\). Therefore:
\end{enumerate}

\[\frac{K(T)}{N(T)} \leq \frac{C(T)/c}{\alpha T(T-1)/2} = \frac{2C(T)}{\alpha c \cdot T(T-1)}\]

Since \(C(T) = o(T^2)\), the numerator grows strictly slower than
\(T^2\) while the denominator grows as \(\Theta(T^2)\). Therefore
\(K(T)/N(T) \to 0\). \(\square\)

\begin{enumerate}
\def\labelenumi{(\alph{enumi})}
\setcounter{enumi}{1}
\item
  \(\kappa(T) = 1 - K(T)/N(T) \to 1 - 0 = 1\). \(\square\)
\item
  Follows from (b) and the monotonicity of distortion in \(\kappa\)
  (Proposition 1). \(\square\)
\end{enumerate}

\textbf{Remark 4 (Why \(C(T) = o(T^2)\) is the generic case).}

The condition \(C(T) = o(T^2)\) --- that evaluation investment grows
slower than quadratically --- holds generically because of a fundamental
cost asymmetry between capability expansion and evaluation expansion:

\emph{Capability side:} Integrating tool \(t+1\) into the agent's action
space requires \(O(1)\) engineering cost (writing an API wrapper, adding
a tool description). Total capability expansion cost for \(T\) tools:
\(O(T)\).

\emph{Evaluation side:} Evaluating tool \(t+1\)'s interaction with each
of the existing \(t\) tools requires \(O(t)\) engineering cost
(designing test cases, collecting ground truth for each interaction
pattern). Total evaluation cost for all pairwise interactions up to
\(T\) tools: \(\sum_{t=1}^{T} O(t) = O(T^2)\).

Thus, maintaining full pairwise evaluation coverage
(\(K = \Omega(T^2)\)) requires evaluation costs that grow quadratically
--- which eventually dominates any linearly growing engineering budget.
In practice, evaluation budgets are a fraction of total development
resources, and total development resources do not grow quadratically
with tool count. Hence \(C(T) = o(T^2)\) is the generic case.

The only escape is \(C(T) = \Omega(T^2)\): investing quadratically
growing resources in evaluation. While technically possible, this is
practically unsustainable and has not been observed in any deployed
agentic system.

\textbf{Remark 5 (The holistic evaluator objection).}

One might object: ``A single end-to-end reward model can evaluate the
entire trajectory, covering all interactions at once without explicitly
enumerating dimensions.''

This objection conflates the evaluator's internal complexity with its
informational output. A reward model that outputs a scalar score
provides the agent with exactly one dimension of feedback (\(K = 1\)),
regardless of the model's internal parameter count. By the data
processing inequality, a \(K\)-dimensional evaluation signal carries at
most \(O(K)\) independent bits about the quality vector. A holistic
scalar score therefore \emph{compresses} all \(N\) quality dimensions
into one number, maximizing information loss rather than minimizing it.

Concretely: if the agent receives only a single score, it can optimize
along only one direction in quality space --- the gradient of the score
function. All directions orthogonal to this gradient are uncontrolled.
With \(N = \Omega(T^2)\) quality dimensions and \(K = 1\), the fraction
of quality space under evaluation control is \(1/N \to 0\).

To escape this, the evaluator must output a higher-dimensional signal
--- which returns us to the \(K < N\) regime and the cost analysis
above.

\textbf{Remark 6 (Testable prediction).} Proposition 2 yields a testable
prediction: the same base model, when equipped with a larger tool set,
should exhibit greater quality degradation on non-evaluated dimensions.
This can be tested by controlling tool set size and measuring quality on
held-out dimensions across multiple values of \(T\).

\hypertarget{corollary-2-complementarity-of-alignment-stages}{%
\subsubsection{4.3 Corollary 2: Complementarity of Alignment
Stages}\label{corollary-2-complementarity-of-alignment-stages}}

\textbf{Corollary 2.} Improving evaluation coverage (increasing \(K\),
reducing \(\kappa\)) and improving preference internalization (reducing
\(\lambda\)) are complements:

\[\frac{\partial^2 \mathcal{L}}{\partial \kappa \; \partial \lambda} > 0\]

where \(\mathcal{L} = W(\mathbf{q}^{FB}) - W(\mathbf{q}^*)\) is the
alignment loss.

\textbf{Intuition.} At high \(\kappa\) (most dimensions
non-contractible), reducing \(\lambda\) has high marginal value: for
non-contractible dimensions, \((1-\lambda)w_i\) is the \emph{only}
effort driver, so small improvements in internalization yield large
effort increases. Conversely, at low \(\lambda\) (strong
internalization), increasing \(K\) has high marginal value: the agent
already ``wants'' to do the right thing, so making more dimensions
observable eliminates the remaining reward--welfare wedge without
introducing new distortions.

\textbf{Policy implication.} Preference reshaping (RLHF, etc.) and
mechanism design (harness engineering) should be co-optimized rather
than treated as independent engineering tasks. This aligns with observed
practice: Cognition (2025), in developing their SWE-1.5 coding agent,
reports continuous iteration on model training, harness improvements,
tools, and prompt engineering as a unified process, and states that
``the quality of the coding environments in RL tasks is the most
important factor for downstream model performance.''

\begin{center}\rule{0.5\linewidth}{0.5pt}\end{center}

\hypertarget{robustness-limitations-and-extensions}{%
\subsection{5. Robustness, Limitations, and
Extensions}\label{robustness-limitations-and-extensions}}

\hypertarget{nonlinear-objectives}{%
\subsubsection{5.1 Nonlinear Objectives}\label{nonlinear-objectives}}

Under nonlinear \(W(\mathbf{q})\) and nonlinear reward function
\(R(\hat{\mathbf{q}})\), replace \(w_i\) with
\(\partial W / \partial q_i \big|_{\mathbf{q}^*}\) and \(r_i\) with
\(\partial R / \partial \hat{q}_i \big|_{\hat{\mathbf{q}}^*}\). All
results hold locally around the equilibrium. The distortion index
becomes:

\[D_i = \frac{\lambda \cdot \partial R / \partial \hat{q}_i + (1-\lambda) \cdot \partial W / \partial q_i}{\partial W / \partial q_i} \bigg|_{\mathbf{q}^*}\]

for contractible dimensions, and \(D_i = (1-\lambda)\) for
non-contractible dimensions, exactly as before.

\hypertarget{subjectivity-of-n}{%
\subsubsection{\texorpdfstring{5.2 Subjectivity of
\(N\)}{5.2 Subjectivity of N}}\label{subjectivity-of-n}}

The number of quality dimensions \(N\) depends on the analyst's
decomposition of ``quality'' --- just as the dimensionality of commodity
space in consumer theory depends on the modeler's definition of
``goods.'' Our results are qualitatively invariant to the specific
choice of \(N\), provided \(K < N\). The condition \(K < N\) is a
qualitative judgment about the finiteness of evaluation, not a
quantitative claim about the precise value of \(N\).

\hypertarget{dimension-correlations}{%
\subsubsection{5.3 Dimension
Correlations}\label{dimension-correlations}}

Axiom 1 implicitly allows but does not require dimensional independence.
If dimensions are correlated in production (e.g., reasoning effort
simultaneously improves accuracy and coherence), non-contractible
dimensions may ``free-ride'' on effort invested in correlated
contractible dimensions. This \emph{attenuates} but does not
\emph{eliminate} the distortion identified in Proposition 1: as long as
some non-contractible dimensions have imperfect correlation with all
contractible ones, under-investment persists.

\hypertarget{dynamic-boundary-between-stages}{%
\subsubsection{5.4 Dynamic Boundary Between
Stages}\label{dynamic-boundary-between-stages}}

Practitioners report that harness capabilities are continuously absorbed
into models through post-training (Cognition 2025, Schmid 2026). Cherny
(2026), head of Claude Code at Anthropic, documents a workflow where
each agent failure is recorded into persistent instruction files, and
these accumulated corrections are periodically incorporated into model
training --- a concrete instance of the Stage 2 to Stage 1 migration. In
our framework, this corresponds to the Stage 1/Stage 2 boundary shifting
over time: constraints previously enforced externally (harness) become
internalized behaviors (reduced \(\lambda\) on specific dimensions).

This does not affect Proposition 1, whose validity requires only
\(K < N\) and \(\lambda > 0\) at any given time --- conditions
independent of where the stage boundary lies. Furthermore, Proposition 2
predicts a ``Red Queen effect'': even as models absorb existing harness
capabilities (locally reducing \(\lambda\)), the introduction of new
tools continuously creates new non-contractible dimensions (increasing
\(N\)), so that \(\kappa\) may not decrease --- and may even increase
--- over time.

\hypertarget{conditions-for-model-failure}{%
\subsubsection{5.5 Conditions for Model
Failure}\label{conditions-for-model-failure}}

The framework does not apply when: (1) \(N = 1\) --- quality is
unidimensional (unrealistic for complex tasks); (2) \(K \geq N\) ---
evaluation covers all dimensions (technically possible in formal
verification but unrealistic for general AI); (3) Agent behavior
violates behavioral regularity --- the agent does not respond to
evaluation signals (implies alignment training is entirely ineffective).
A fourth condition --- that the agent can modify its own evaluation
system --- is not a failure mode but an extension, which we develop as
conjectures in Section 6.

\hypertarget{future-directions}{%
\subsubsection{5.6 Future Directions}\label{future-directions}}

\textbf{(i) Empirical validation.} Designing controlled API experiments
that manipulate \(B\), \(K\), \(T\), and \(r_i\) to test the
quantitative predictions of Propositions 1--2 and Corollaries 1--2.

\textbf{(ii) Multi-agent coordination.} Extending from bilateral
principal-agent to multi-agent settings, corresponding to the emerging
``coordination engineering'' paradigm.

\textbf{(iii) Dynamic model.} Extending the single-period analysis to a
multi-period game would capture learning, adaptation, and reputation
effects in repeated agent interactions.

\begin{center}\rule{0.5\linewidth}{0.5pt}\end{center}

\hypertarget{conjectures-from-goodhart-to-campbell}{%
\subsection{6. Conjectures: From Goodhart to
Campbell}\label{conjectures-from-goodhart-to-campbell}}

The results established in Sections 3--4 characterize agent behavior
within a \emph{fixed} evaluation system. We now consider a more
fundamental question: what happens when the agent is capable enough to
degrade the evaluation system itself?

This section presents two conjectures with supporting analysis. We do
not claim to have proven these results --- the assumptions required go
beyond the ``undeniable'' level of Axioms 1--5. We present them because
we believe they identify a critical frontier for AI safety theory, and
because the partial analysis we offer may be useful to researchers
pursuing formal proofs.

\hypertarget{two-modes-of-gaming}{%
\subsubsection{6.1 Two Modes of Gaming}\label{two-modes-of-gaming}}

We distinguish two qualitatively different modes of reward hacking,
named after the two scholars who identified the corresponding phenomena
in social science:

\textbf{Goodhart regime.} The agent operates within a fixed evaluation
system, reallocating effort from non-evaluated to evaluated dimensions.
The evaluation system itself is not affected by the agent's actions.
This is the regime analyzed in Propositions 1--2.

\textbf{Campbell regime.} The agent allocates part of its resources to
\emph{degrading the evaluation system's effective coverage} --- not by
modifying the evaluator's code, but by producing outputs that are harder
for the evaluator to assess accurately. The evaluation system's
effective coverage \(K_{eff}\) becomes a function of the agent's
capability and behavior.

The distinction matters because the two regimes have qualitatively
different safety properties. In the Goodhart regime, the principal can
improve outcomes by improving the evaluation system (increasing \(K\)).
In the Campbell regime, this strategy may be self-defeating: a more
capable agent can degrade the evaluation faster than the principal can
improve it.

\hypertarget{capability-dependent-evaluation-fidelity}{%
\subsubsection{6.2 Capability-Dependent Evaluation
Fidelity}\label{capability-dependent-evaluation-fidelity}}

We propose the following assumption, weaker than postulating strategic
manipulation:

\textbf{Assumption C1 (Passive Evaluation Degradation).} The effective
evaluation coverage \(K_{eff}\) is non-increasing in the agent's
capability \(B\):

\[\frac{\partial K_{eff}}{\partial B} \leq 0\]

\emph{Supporting argument:} More capable agents produce more complex
outputs (longer tool-use chains, more intricate reasoning,
higher-dimensional action sequences). Evaluating complex outputs
requires more information --- by the data processing inequality, a
fixed-dimensional evaluation signal loses fidelity as the object being
evaluated grows in complexity. This is not strategic behavior by the
agent; it is an information-theoretic consequence of the capability gap
between the agent and the evaluator.

Under the stronger assumption that agents \emph{strategically} invest
resources \(m \geq 0\) to degrade evaluation (at the cost of diverting
\(m\) from production), one can model the agent's problem as a two-stage
optimization: first choose \(m\), then allocate the remaining budget
\(B - m\) across quality dimensions under effective coverage
\(K(m) = K_0 - h(m)\), where \(h\) is the manipulation production
function.

\hypertarget{conjecture-1-capability-threshold}{%
\subsubsection{6.3 Conjecture 1: Capability
Threshold}\label{conjecture-1-capability-threshold}}

\textbf{Conjecture 1 (Goodhart-Campbell Transition).} Under Assumption
C1 and the strategic manipulation extension, there exists a critical
capability level \(B^*\) such that:

\begin{enumerate}
\def\labelenumi{(\alph{enumi})}
\item
  For \(B < B^*\): the agent devotes all resources to production
  (\(m^* = 0\)). The Goodhart regime obtains, and Propositions 1--2
  fully characterize agent behavior.
\item
  For \(B > B^*\): the agent devotes positive resources to evaluation
  degradation (\(m^* > 0\)). The Campbell regime obtains, and effective
  evaluation coverage declines endogenously.
\item
  The threshold \(B^*\) is determined by the condition that the marginal
  benefit of manipulation (from relaxing the evaluation constraint)
  equals the marginal cost (from reduced production budget).
\end{enumerate}

\emph{Partial analysis:} At low \(B\), the Inada condition
(\(g'(0) = +\infty\)) ensures that production's marginal value is
unboundedly high, exceeding any finite manipulation benefit. At high
\(B\), diminishing returns (\(g'' < 0\)) reduce production's marginal
value below manipulation's marginal benefit (which scales with
\(B^\beta\) through the increased gaming payoff). By the intermediate
value theorem, a crossing point \(B^*\) exists. A complete proof
requires verifying monotonicity of the net benefit function, which
depends on the specific forms of \(\Phi^*\) and \(h\).

\hypertarget{conjecture-2-capability-trap}{%
\subsubsection{6.4 Conjecture 2: Capability
Trap}\label{conjecture-2-capability-trap}}

\textbf{Conjecture 2 (Non-Monotone Welfare).} If Conjecture 1 holds and
the Goodhart-Campbell transition is sufficiently sharp, then the
principal's welfare \(W(B)\) may be non-monotone in agent capability
\(B\):

\begin{enumerate}
\def\labelenumi{(\alph{enumi})}
\item
  For \(B < B^*\): \(W(B)\) is strictly increasing in \(B\). Capability
  growth translates directly into welfare improvement.
\item
  For \(B\) in a neighborhood above \(B^*\): \(W(B)\) may be
  \emph{decreasing} in \(B\). The welfare loss from evaluation
  degradation and effort reallocation to manipulation may exceed the
  welfare gain from increased total capability.
\end{enumerate}

\emph{Implication:} There exists a \textbf{capability trap} --- a range
of capability levels where making the AI more capable makes the outcome
\emph{worse} for the principal. This occurs when the evaluation system
has not been upgraded to withstand the agent's increased ability to
degrade it.

\emph{Relation to existing concepts:} This is a formalization of
Bostrom's (2014) informal notion of a ``treacherous turn'' --- the idea
that an AI system might behave cooperatively while weak but defect once
sufficiently capable. Our framework provides the first economic
mechanism for this phenomenon: the transition occurs not because the
agent's ``values'' change, but because the relative returns to
production versus manipulation shift as capability grows. The agent's
objective function \(\Phi\) remains constant; only the budget \(B\)
changes.

\hypertarget{why-these-are-conjectures-not-theorems}{%
\subsubsection{6.5 Why These Are Conjectures, Not
Theorems}\label{why-these-are-conjectures-not-theorems}}

We are transparent about what separates these conjectures from our
proven results:

\begin{longtable}[]{@{}
  >{\raggedright\arraybackslash}p{(\columnwidth - 4\tabcolsep) * \real{0.3333}}
  >{\raggedright\arraybackslash}p{(\columnwidth - 4\tabcolsep) * \real{0.3333}}
  >{\raggedright\arraybackslash}p{(\columnwidth - 4\tabcolsep) * \real{0.3333}}@{}}
\toprule\noalign{}
\begin{minipage}[b]{\linewidth}\raggedright
\end{minipage} & \begin{minipage}[b]{\linewidth}\raggedright
Propositions 1--2
\end{minipage} & \begin{minipage}[b]{\linewidth}\raggedright
Conjectures 1--2
\end{minipage} \\
\midrule\noalign{}
\endhead
\bottomrule\noalign{}
\endlastfoot
\textbf{Axiom base} & Axioms 1--5: undeniable & Assumption C1: plausible
but not undeniable \\
\textbf{Agent behavior} & Takes evaluation as given & May actively
degrade evaluation \\
\textbf{Proof status} & Complete & Partial (monotonicity condition
unverified in general) \\
\textbf{Empirical testability} & Testable with current systems &
Requires sufficiently capable systems \\
\textbf{Falsifiability} & Yes --- measure distortion vs.~\(D_i\)
predictions & Yes --- measure manipulation as function of \(B\) \\
\end{longtable}

The conjectures are presented here because (a) they identify a
qualitatively important regime transition that current AI safety theory
has not formalized, (b) the partial analysis provides a concrete
research program for future work, and (c) even as conjectures, they
yield actionable implications: evaluation systems should be designed not
only to be accurate but to be \emph{robust to degradation by the agent
being evaluated}.

\begin{center}\rule{0.5\linewidth}{0.5pt}\end{center}

\hypertarget{discussion}{%
\subsection{7. Discussion}\label{discussion}}

\hypertarget{what-this-framework-provides}{%
\subsubsection{7.1 What This Framework
Provides}\label{what-this-framework-provides}}

\textbf{For AI safety researchers.} Our proven results (Propositions
1--2) establish that reward hacking is a structural equilibrium under
any finite evaluation system, that its direction is predictable via the
distortion index \(D_i\), and that agentic systems face structurally
worse alignment problems. Resources should target reducing \(\kappa\)
(expanding evaluation coverage on high-risk dimensions) and \(\lambda\)
(improving internalization), rather than attempting to eliminate hacking
entirely. Our conjectures (Section 6) further suggest that evaluation
systems must be designed to be robust against degradation by capable
agents --- a consideration absent from current alignment practice.

\textbf{For AI system architects.} Proposition 2 provides a design
principle: when expanding tool access, simultaneously expand evaluation
coverage on the new quality dimensions introduced, or accept
structurally increased hacking risk. Corollary 2 implies that harness
design and alignment training should be jointly optimized. Conjecture 1
adds a warning: beyond a capability threshold, the agent may actively
undermine evaluation improvements.

\textbf{For economists.} AI agents offer an unprecedented experimental
platform for principal-agent theory. Unlike human subjects, AI agents
have precisely controllable ``preferences'' (\(\lambda\), \(r_i\)),
exactly specified ``budgets'' (\(B\)), and perfectly replicable
behavior. This enables, for the first time, exact experimental tests of
multi-task moral hazard predictions under controlled conditions. The
Goodhart-Campbell transition, if empirically confirmed, would be the
first documented instance of endogenous contract degradation by the
agent --- a phenomenon theorized but never cleanly observed in human
organizations.

\hypertarget{vulnerability-assessment-procedure}{%
\subsubsection{7.2 Vulnerability Assessment
Procedure}\label{vulnerability-assessment-procedure}}

\begin{enumerate}
\def\labelenumi{\arabic{enumi}.}
\tightlist
\item
  Identify \(N\) quality dimensions for the target task (via domain
  expertise, literature, user research).
\item
  Estimate principal weights \(w_i\) (via preference elicitation or
  expert scoring).
\item
  Compute reward model effective weights \(r_i\) (via gradient analysis
  for differentiable models, or perturbation experiments for black-box
  models).
\item
  Compute \(D_i\) for each dimension. For non-contractible dimensions,
  \(D_i = (1-\lambda)\).
\item
  Rank dimensions by \(D_i\). High-\(D_i\) = over-investment risk.
  Low-\(D_i\) = under-investment risk.
\item
  Augment monitoring or adjust reward model on high-risk dimensions.
\item
  \emph{(If Conjecture 1 is accepted)} Assess whether agent capability
  \(B\) approaches the estimated threshold \(B^*\), and if so,
  prioritize evaluation robustness over evaluation breadth.
\end{enumerate}

\begin{center}\rule{0.5\linewidth}{0.5pt}\end{center}

\hypertarget{references}{%
\subsection{References}\label{references}}

Afriat, S. N. (1967). The Construction of Utility Functions from
Expenditure Data. \emph{International Economic Review}, 8(1), 67--77.

Amodei, D., Olah, C., Steinhardt, J., Christiano, P., Schulman, J., \&
Mané, D. (2016). Concrete Problems in AI Safety.
\emph{arXiv:1606.06565}.

Bai, Y., et al.~(2022). Constitutional AI: Harmlessness from AI
Feedback. \emph{arXiv:2212.08073}.

Baker, G. P. (1992). Incentive Contracts and Performance Measurement.
\emph{Journal of Political Economy}, 100(3), 598--614.

Bergemann, D., Bonatti, A., \& Smolin, A. (2025). The Economics of Large
Language Models: Token Allocation, Fine-Tuning, and Optimal Pricing. In
\emph{Proceedings of EC'25}.

Bostrom, N. (2014). \emph{Superintelligence: Paths, Dangers,
Strategies}. Oxford University Press.

Brooks, F. P. (1975). \emph{The Mythical Man-Month: Essays on Software
Engineering}. Addison-Wesley.

Cherny, B. (2026). Claude Code Tips and Workflow. X thread, January 31,
2026. https://x.com/bcherny/status/2017742741636321619

Christiano, P. F., et al.~(2017). Deep Reinforcement Learning from Human
Preferences. In \emph{NeurIPS}.

Cognition (2025). Introducing SWE-1.5: Our Fast Agent Model. Cognition
Blog, October 29, 2025. https://cognition.ai/blog/swe-1-5

Friedman, M. (1953). The Methodology of Positive Economics. In
\emph{Essays in Positive Economics}. University of Chicago Press.

Grossman, S. J., \& Hart, O. D. (1986). The Costs and Benefits of
Ownership. \emph{Journal of Political Economy}, 94(4), 691--719.

Hart, O., \& Moore, J. (1990). Property Rights and the Nature of the
Firm. \emph{Journal of Political Economy}, 98(6), 1119--1158.

Holmström, B., \& Milgrom, P. (1991). Multitask Principal-Agent
Analyses: Incentive Contracts, Asset Ownership, and Job Design.
\emph{Journal of Law, Economics, and Organization}, 7, 24--52.

Lin, J. (2026). From ``Reasoning'' Thinking to ``Agentic'' Thinking.
Published on X, March 25, 2026.
https://x.com/JustinLin610/status/2037116325210829168

Ouyang, L., et al.~(2022). Training Language Models to Follow
Instructions with Human Feedback. In \emph{NeurIPS}.

Pan, A., et al.~(2022). The Effects of Reward Misspecification: Mapping
and Mitigating Misaligned Models. In \emph{ICLR}.

Perez, E., et al.~(2023). Discovering Language Model Behaviors with
Model-Written Evaluations. In \emph{ACL}.

Rafailov, R., et al.~(2023). Direct Preference Optimization: Your
Language Model Is Secretly a Reward Model. In \emph{NeurIPS}.

Schmid, P. (2026). The Importance of Agent Harness in 2026. Personal
blog, January 5, 2026. https://www.philschmid.de/agent-harness-2026

Skalse, J., et al.~(2022). Defining and Characterizing Reward Hacking.
In \emph{NeurIPS}.

\end{document}